\documentclass[conference]{IEEEtran}

\IEEEoverridecommandlockouts

\usepackage{cite}
\usepackage{xcolor}
\usepackage{acronym}
\usepackage{graphicx}
\usepackage{textcomp}
\usepackage{hyperref}
\usepackage{algorithmic}
\usepackage{amsmath,amssymb,amsfonts}

\acrodef{AP}{Average Precision}
\acrodef{YOLO}{You Only Look Once}
\acrodef{mAP}{mean Average Precision}
\acrodef{IoU}{Intersection over Union}
\acrodef{SPP}{Spatial Pyramid Pooling}
\acrodef{GPU}{Graphics Processing Unit}
\acrodef{ANN}{Artificial Neural Network}
\acrodef{PANet}{Path Aggregation Network}
\acrodef{CSP}{Center and Scale Prediction}
\acrodef{SURF}{Speeded up Robust Features}
\acrodef{BDL}{Boosting-like Deep Learning}
\acrodef{CNN}{Convolutional Neural Network}
\acrodef{RRC}{Recurrent Rolling Convolution}
\acrodef{CSPNet}{Cross-Stage Partial Network}
\acrodef{HOG}{Histogram of Oriented Gradients}
\acrodef{Faster R-CNN}{Faster Region-based CNN}
\acrodef{SIFT}{Scale-invariant Feature Transform}
\acrodef{ADAS}{Advanced Driver-Assistance Systems}
\acrodef{FLOPS}{Floating-point Operations-per-Second}

\newcommand{\wrt}{w.r.t. }
\newcommand{\yolo}{YOLOv5}
\newcommand{\yolos}{YOLOv5s}

\newcommand{\eg}{\textit{e.g., }}
\newcommand{\ie}{\textit{i.e., }}
\newcommand{\etal}{\textit{et al. }}
\newcommand{\caltech}{Caltech pedestrian dataset}

\def\BibTeX{{\rm B\kern-.05em{\sc i\kern-.025em b}\kern-.08em
    T\kern-.1667em\lower.7ex\hbox{E}\kern-.125emX}}
\begin{document}

\title{A Robust Pedestrian Detection Approach for Autonomous Vehicles
}

\author{
    \IEEEauthorblockN{Bahareh Ghari}
    \IEEEauthorblockA{\textit{Department of Computer Engineering} \\
    \textit{University of Guilan} \\
    Rasht, Iran \\
    baharehghari@msc.guilan.ac.ir}
    \and
    \IEEEauthorblockN{Ali Tourani}
    \IEEEauthorblockA{\textit{Interdisciplinary Centre for Security,} \\ \textit{Reliability, and Trust (SnT)} \\
    \textit{University of Luxembourg}\\
    Luxembourg, Luxembourg \\
    ali.tourani@uni.lu}
    \and
    \IEEEauthorblockN{Asadollah Shahbahrami}
    \IEEEauthorblockA{\textit{Department of Computer Engineering} \\
    \textit{University of Guilan} \\
    Rasht, Iran \\
    shahbahrami@guilan.ac.ir}
}

\maketitle

\begin{abstract}
Nowadays, utilizing \ac{ADAS} has absorbed a huge interest as a potential solution for reducing road traffic issues.
Despite recent technological advances in such systems, there are still many inquiries that need to be overcome.
For instance, \ac{ADAS} requires accurate and real-time detection of pedestrians in various driving scenarios.
To solve the mentioned problem, this paper aims to fine-tune the \yolos~framework for handling pedestrian detection challenges on the real-world instances of \caltech. 
We also introduce a developed toolbox for preparing training and test data and annotations of \caltech~into the format recognizable by \yolo.
Experimental results of utilizing our approach show that the \ac{mAP} of our fine-tuned model for pedestrian detection task is more than $91$ percent when performing at the highest rate of ~70 FPS.
Moreover, the experiments on the \caltech~samples have verified that our proposed approach is an effective and accurate method for pedestrian detection and can outperform other existing methodologies.
\end{abstract}


\section{Introduction}
\label{sec_intro}

As a well-known issue, many people get injured or lose their lives in traffic accidents every day.
There are several causes for these accidents, such as driver tiredness, speeding, and drunk driving.
Therefore, autonomous systems have been utilized in modern vehicles to provide reliable and safe driving and decease the risk of accidents \cite{adnan2018trust}.
These systems use various sensors to analyze the surroundings in real-time and provide automatic decision-making while driving.
In this regard, pedestrian detection is a vital component in automotive safety and \acf{ADAS} domains, as it can guarantee safe vehicle passage on the roads \cite{zhang2017towards}.

The process of pedestrian detection in autonomous vehicles can lie under object detection tasks.
Object detection has recently matured and overwhelmed any other hand-craft or classic detection solutions \cite{hnewa2020object}.
Currently, there is a vast number of object detection algorithms, including \ac{HOG} \cite{dalal2005histograms}, \ac{SIFT} \cite{lowe2004distinctive}, \ac{SURF} \cite{bay2006surf}, Haar feature-based methods \cite{viola2001rapid}, and \acp{ANN} \cite{da2017artificial}.
Object detection contributes to the \acp{CNN} \cite{gu2018recent} in the computer vision domain, providing robust outcomes compared to the classic image processing approaches \cite{jiao2019survey}.
In this regard, many deep learning-based applications for object detection are divided into either "one-stage" -\textit{\eg \ac{YOLO} \cite{redmon2016you}}- or "two-stage" -\textit{\eg \ac{Faster R-CNN} \cite{ren2015faster}}- methodologies.
While in the two-stage methods a set of samples is used to find the candidate objects, one-stage techniques predict the locations as a regression problem.

\begin{figure}[t!]
    \centering
    \includegraphics[width=\linewidth]{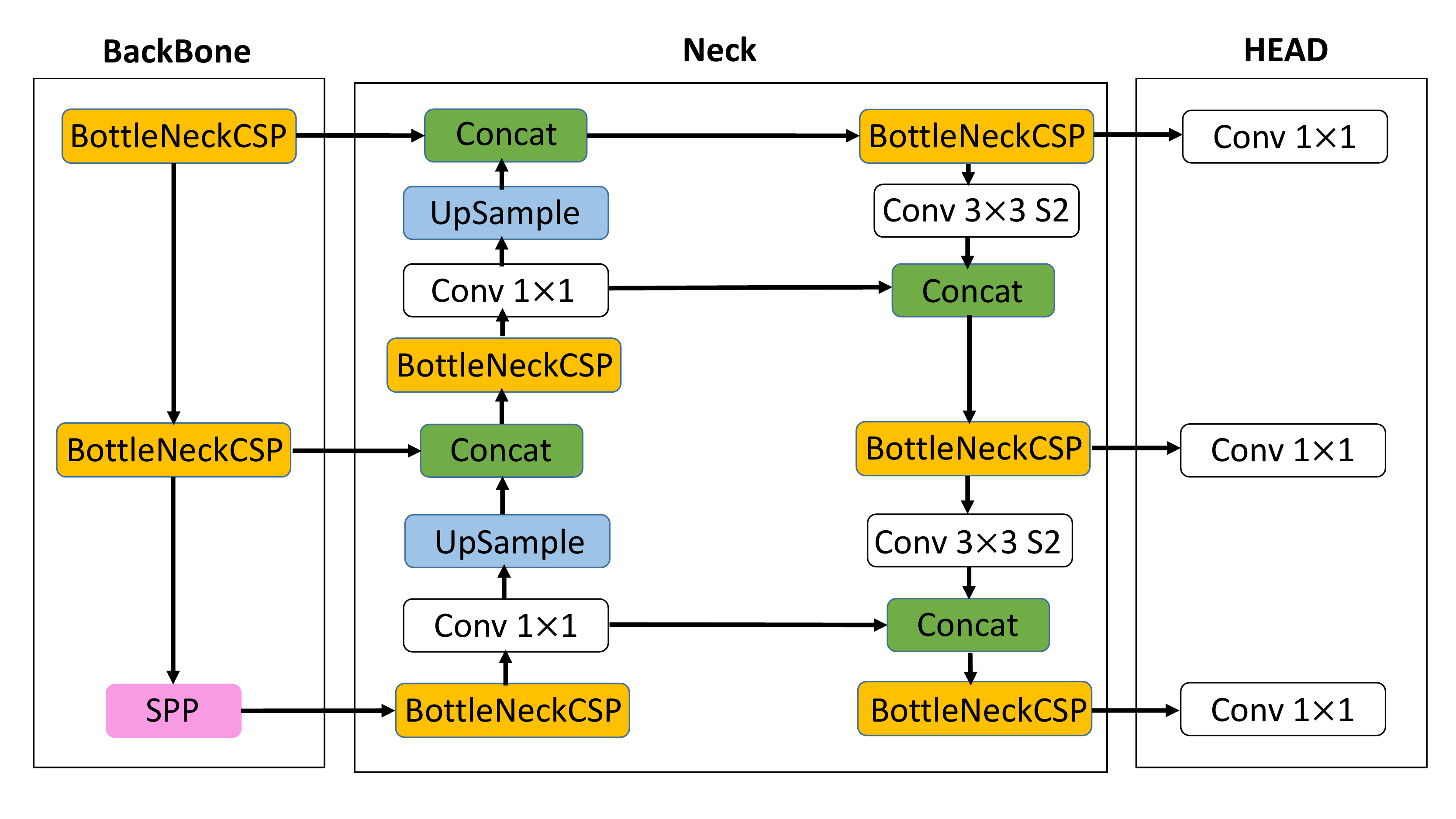}
    \caption{The network architecture of \yolo, which includes three modules: (1) \textit{Backbone}: CSPDarknet, (2) \textit{Neck}:  PANet, and (3) \textit{Head}: \ac{YOLO}v3 (anchor-based). The input data includes mosaic data augmentation and anchor box calculation are first input to \textit{CSPDarknet} for feature extraction, and then fed to \textit{PANet} to boost information flow. Finally, the \textit{Head} consumes features from the \textit{Neck} and outputs the results such as bounding boxes and class predictions. Hereby, \textit{SPP} stands for Spatial Pyramid Pooling and \textit{Concat} refers to concatenation.}
    \label{fig_architecture}
\end{figure}

Among a wide range of algorithms, \ac{YOLO} is a one-stage object detector that utilizes a real-time end-to-end approach to predict bounding boxes and class labels at once \cite{redmon2016you}.
At the time of writing this paper, many different versions and variations have been introduced.
\ac{YOLO}v2 has improved the performance of the first version, including low recall and small object detection, by using batch normalization on convolutional layers, improved classifier, and anchor boxes \cite{redmon2017yolo9000}.
As the fast approach introduced in \ac{YOLO}v2 decreased the accuracy, \ac{YOLO}v3 appeared with a robust backbone, allowing it to detect features at three scales \cite{redmon2018yolov3}.
Although later versions of \ac{YOLO} are not considered official, their developers have changed the architecture of \ac{YOLO}v3 to improve performance and accuracy.
In \ac{YOLO}v4 \cite{bochkovskiy2020yolov4}, the primary focus is on enhancing the object detection stage, resulting in ten and twelve percent gains in accuracy and performance, respectively.
In $2021$, the fifth version was introduced as compound-scaled object detection models trained on the COCO dataset \cite{lin2014microsoft}.
\yolo \cite{jocher2021ultralytics} is implemented with the aid of the \textit{Ultralytics PyTorch} framework and Python programming language, making it a super fast methodology to train.

In this paper, a pedestrian detection application based on \yolo~is introduced.
The main contributions of the paper are described below:

\begin{itemize}
    \item Fine-tuning of \yolo~for handling the pedestrian detection challenges,
    \item Implementing a toolbox for preparing \caltech~training and test data and annotations into proper \yolo~format.
\end{itemize}

The rest of the paper is organized as follows: Section \ref{sec_related} reviews recent pedestrian detection approaches.
In Section \ref{sec_proposed}, we discuss our proposed method and contributions for pedestrian detection in detail.
Section \ref{sec_evaluation} presents the experimental results and discussions, and finally, the paper concludes in Section \ref{sec_conclude}.
\section{Related Works}
\label{sec_related}

Many recent publications have recommended deep learning-based solutions for pedestrian detection and tracking.
Most of these researches have focused on the overall architecture of the detector module for a reliable pedestrian detection and tracking.
For instance, Boyuan \etal \cite{boyuan2020study} have proposed a detection model that combines a new type of \ac{SPP} network and K-means clustering algorithm with \ac{YOLO}v4 model for easier feature extraction. In \cite{tian2015deep}, \textit{DeepParts} is proposed, consisting of forty-five fine-tuned part detectors to tackle the occlusion challenge.
Each part detector can cover the entire body of pedestrians at different positions and scales with partial occlusion.
Similarly, Noh \etal \cite{noh2018improving} adopted a set of part-based detectors learned and integrated into single-stage networks to solve the occlusion problem.
For domain adaptation, Zeng \etal \cite{zeng2014deep} have proposed a new approach to automatically learn domain invariant features and visual patterns in static video surveillance with no manual labels from the target scene.
In \cite{li2019deep}, weighted combination layer and depth-wise separable convolution were combined with \ac{YOLO} to improve the accuracy and speed of detecting pedestrians in hazy weather.

Moreover, based on the recent advances in the anchor-free detection networks, \ac{CSP} \cite{liu2019high} is proposed to detect the center and scale of pedestrians with a fixed aspect ratio.
In \cite{ren2017accurate}, \ac{RRC} architecture is introduced to gradually aggregate relevant contextual information among the multi-scale feature maps.
Based on the channel-wise attention mechanism, Zhang \etal \cite{zhang2018occluded} added an attention network to the \ac{Faster R-CNN} architecture to handle different occlusion patterns in pedestrian detection.
Some other studies addressed the role of data and tuning in pedestrian detection applications.
For instance, the \ac{BDL} framework \cite{wang2015boosting} is used to prevent the over-fitting problem while training by reducing the average miss rate of the detector.
Additionally, and as a baseline, Tumas \etal \cite{tumas2020pedestrian} introduced a dataset for detecting pedestrians in challenging weather conditions.

Unlike the mentioned studies, our method fine-tunes the robust architecture of \yolo~object detector for the pedestrian detection task.
The fundamental goal of this paper is to provide accurate pedestrian detection in different scenarios, such as profound environmental changes, while keeping the architecture simple and fast.
We also aim to provide detection in real-time so that our approach can be used in driverless vehicles.
\section{Proposed Method}
\label{sec_proposed}

Due to the splendid features introduced in \yolo, including a lower training time, auto-learning anchor bounding boxes, and mosaic data augmentation, we have utilized this version in our approach.

\subsection{Architecture}
\label{subsec_architecture}

The network structure of \yolo~consists of three primary parts, as shown in \figurename{\ref{fig_architecture}}.
Accordingly, the backbone is a \ac{CNN} that aggregates and forms image features at different granularities.
\yolo~adopts \acp{CSPNet} \cite{wang2020cspnet} as its backbone to formulate image features.
The \textit{CSPNet} addresses duplicate gradient problems in deeper \acp{CNN}, resulting in fewer parameters and \ac{FLOPS} of the model, improving the inference speed and accuracy and reducing the model size.
The network also adds an \ac{SPP} block \cite{he2015spatial} after the \textit{CSP} to remove the fixed-size input image constraint.
The \textit{\ac{SPP}} block is used to compute the feature maps from the entire image only once, and then pool features in arbitrary regions (sub-images) to increase the receptive field and generate fixed-length representations for training the detectors.
The \textit{Neck} in the model is a series of layers to combine image features to pass them forward for the detection stage.
It employs a \ac{PANet} \cite{liu2018path} to boost the information flow process.

Specifically, the feature pyramid is enhanced with accurate localization signals in lower layers by bottom-up path augmentation, which shortens the information path between the lower layers and the top feature.
At the same time, adaptive feature pooling is employed, which links the feature grid and all feature levels to propagate helpful information in each level directly to the following proposal sub-networks.
The \textit{Head} of the model is mainly used to perform the final detection part.
It applied feature anchor boxes and generated final output vectors with class probabilities, objectness scores, and bounding boxes regression.
\yolo~implements the same \ac{YOLO}v3 (anchor-based) head for prediction. In addition \yolo including four models ranging from small to large according to the memory storage size (parameters): \yolos~(the smallest, which we have used in this paper), \ac{YOLO}v5m (medium), \ac{YOLO}vl (large), and \ac{YOLO}v5x (extra large, the most prominent).
All four models were trained on the MS COCO \cite{lin2014microsoft} training dataset.

\subsection{Dataset}
\label{subsec_dataset}

\begin{figure}[t!]
    \centering
    \includegraphics[width=\linewidth]{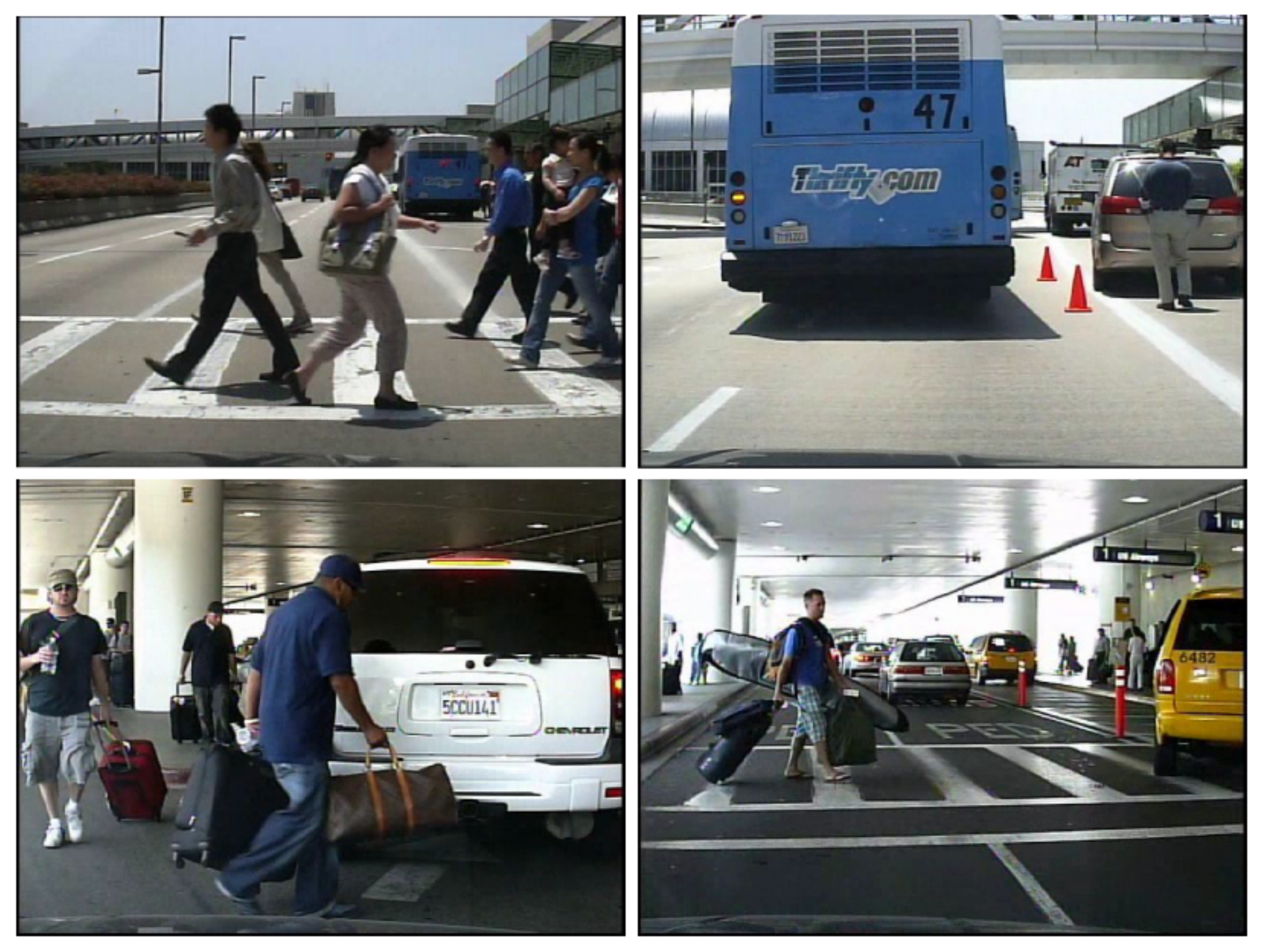}
    \caption{Some instances of the \caltech, the baseline for training and testing in this paper.}
    \label{fig_dataset}
\end{figure}

We have utilized the \caltech \cite{dollar2011pedestrian} in this work, which contains more than ten hours of real-world videos taken from a vehicle driving through regular traffic in an urban environment.
The resolution of the videos is \(640\times480\) and the total number of frames exceeds $250,000$.
Also, challenges in the \caltech~ such as variation in lighting conditions, pedestrian size, and occlusion, making this dataset a proper choice for the training of pedestrian detectors.
This dataset contains more than $350,000$ bounding boxes and $2,300$ unique pedestrians annotated.
\figurename{\ref{fig_dataset}} shows some instances of the dataset.

\subsection{Data preparation for \yolo}
\label{subsec_prepare}

As \ac{YOLO} does not recognize the Caltech's training data and annotations, we needed to implement a conversion toolbox for it.
The mentioned toolbox contains different tools to prepare proper feed for \ac{YOLO} architecture and is publicly accessible in a GitHub repository
\footnote{\url{https://github.com/GuilanITS/Caltech-Pedestrian-YOLO}}.
Using our toolbox, we processed \caltech training \textit{.seq} files and generated squared \(640\times640\) \textit{.png} images.
Additionally, the \textit{.vbb} annotation files are converted into \textit{.txt} files according to the \yolo~standard.

\subsection{Training Process}
\label{subsec_training}

We picked the most miniature version of the \yolo~framework for the experiment, titled \yolos.
After fine-tuning the architecture of \yolos~based on the application requirements, the network has been trained on $3,000$ instances of the dataset in $600$ epochs with a learning rate of $0.01$, a momentum of $0.937$ and a batch size of eight. 
The number of images in the validation and test sets was $280$ and $1006$, respectively.
The total number of parameters in the network was $7,012,822$, shaped into $213$ layers.
Pre-trained weights on COCO dataset are used to initialize the detection model.
Among SGD, Adam and AdamW optimization functions implemented in \yolo, we chose the Adam algorithm for the optimization of parameters.
With this settings, $600$ epochs completed in $10.902$ hours.
Moreover, \yolo~ can easily trade-off between speed and accuracy by changing the size of input image, without retraining.
For instance, at the resolution of $1280$, \yolos~runs in $4.9ms$ at $83.3$ \ac{mAP}.
So, larger image sizes usually lead to better results, but obviously, take longer to process.
It should also be noted that we used only \textit{person} class in the dataset and ignored the \textit{people} class, enabling us to detect individuals correctly even if they are in a crowd.
\section{Evaluation}
\label{sec_evaluation}

The machine for conducting experiments was equipped with an \textit{NVIDIA GeForce RTX3090} \ac{GPU} with $24,576$ MegaBytes of memory and an \textit{Intel(R) Xeon(R) Gold 6248R} processor.
The codes were implemented in \textit{Python v3}, and we used \textit{CUDA v.11.6.55} computing platform and \textit{PyTorch} framework.

\begin{figure}[t!]
    \centering
    \includegraphics[width=\linewidth]{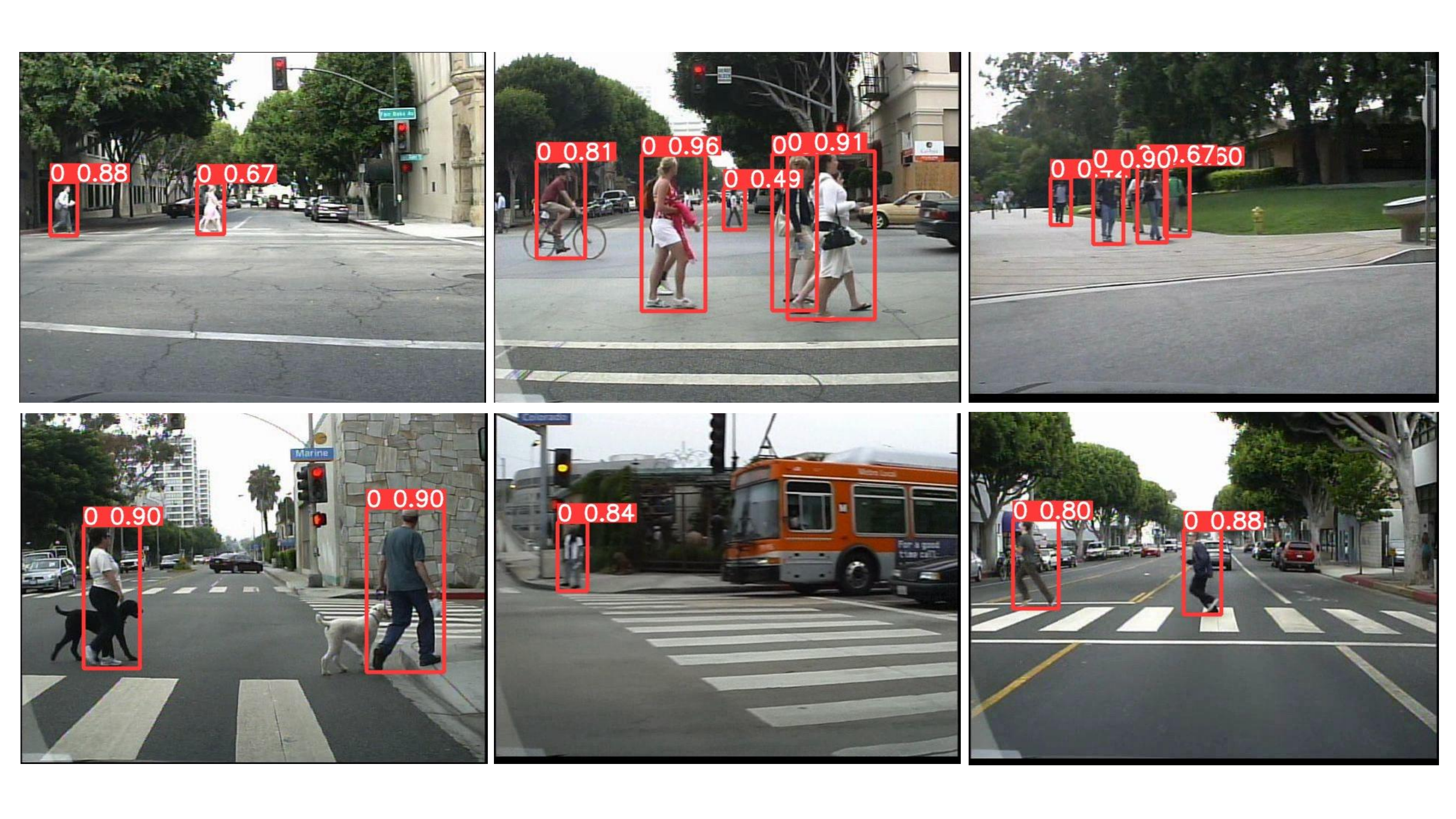}
    \caption{Detection results using the fine-tuned \yolos~model proposed in this paper.}
    \label{fig_experiments}
\end{figure}

Fig. \ref{fig_experiments} demonstrates the pedestrian detection results of the proposed approach in various scenarios.
We can see that even people walking in the throngs have been detected and tracked correctly using our fine-tuned network.

\begin{figure}[t!]
    \centering
    \includegraphics[width=\linewidth]{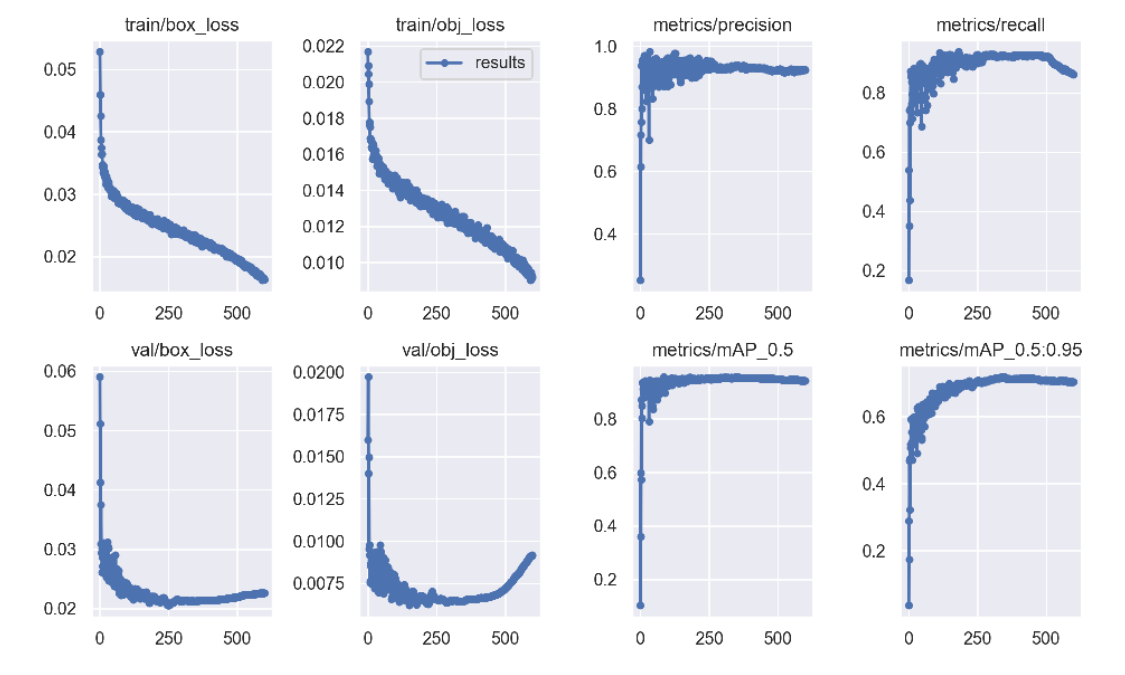}
    \caption{Fluctuations of key indicators while training in different epochs. The top five curves are related to network training and the bottom five curves are related to network validation. In all graphs, the x-axis indicates the epochs. The y-axis refers to the loss or accuracy values, based on the graph.}
    \label{fig_curves}
\end{figure}

\subsection{Evaluation Metrics}
\label{subsec_evalmetrics}

We have employed precision, recall, \ac{AP}, and \ac{mAP} metrics to evaluate our proposed method.
Precision refers to the ratio of all correctly predicted instances (\ie pedestrians) among all predictions.
Recall is used to indicate the number of correctly classified samples in the total number of ground-truth data.
In other words, it shows how many sample was the model able to detect out of the total number of pedestrians in the input image.
The mentioned metrics are calculated using the equations \ref{equation_pr} and \ref{equation_rc}, respectively.

\begin{equation}
\label{equation_pr}
    precision = \frac{TP}{TP + FP}
\end{equation}

\begin{equation}
\label{equation_rc}
    recall = \frac{TP}{TP + FN}
\end{equation}

Where \textit{TP} is True-Positive, \textit{FP} refers to False-Positive, and \textit{FN} refers to False-Negative.
A True-Positive is an outcome where the model correctly predicts the positive class.
A False-Positive is an outcome where the model incorrectly predicts the positive class.
And similarly, a False-Negative is an outcome where the model incorrectly predicts the negative class.
The value of precision and recall depends on how many True-Positives were detected by the model.
A precision-recall curve plots the value of precision against recall for different confidence threshold values.
To simplify using precision and recall measure, we use F1-score, which shapes a single metric based on their harmonic mean.
Equation \ref{equation_f1} shows calculation of F1-score, where \textit{Pr} and \textit{Rc} refer to precision and recall, respectively.

\begin{equation}
\label{equation_f1}
    F1-score = 2\times \frac{(Pr \times Rc)}{Pr + Rc}
\end{equation}

\ac{AP} represents the area under the precision-recall curve and summarizes the precision-recall curve to one scalar value.
The range for AP is between 0 to 1.
The metric \ac{AP} can be calculated from Equation~\ref{equation_ap}, where \textit{n} is the number of thresholds.

\begin{equation}
\label{equation_ap}
    AP = \sum_{k=0}^{k=n-1} (Rc(k) - Rc(k+1)) \times Pr(k)
\end{equation}

Additionally, \ac{mAP} is another metric to evaluate object detection models indicating the average \ac{AP} for each category. 
The formula for calculating \ac{mAP} is shown in Equation \ref{equation_map}, in which \textit{k} refers to a specific class and \textit{n} refers to the total number of classes.

\begin{equation}
\label{equation_map}
    mAP = \frac{1}{n}\sum_{k=1}^{k=n} AP_{k}
\end{equation}

\subsection{Evaluation Results}
\label{subsec_evalresults}

\begin{figure}[t!]
    \centering
    \includegraphics[width=\linewidth]{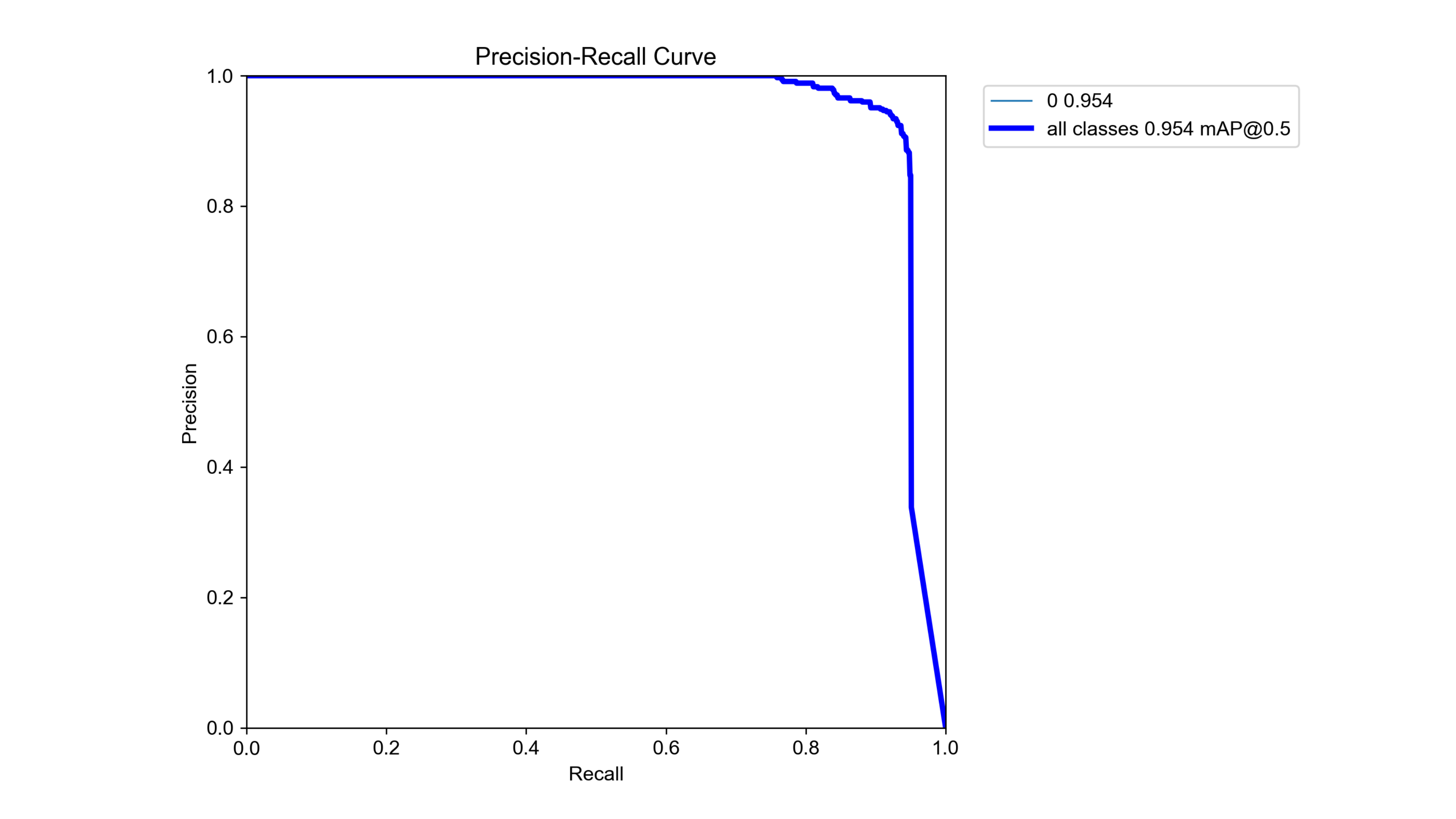}
    \caption{The Precision-Recall curve of the proposed method.}
    \label{fig_prcurve}
\end{figure}

\begin{table*}[!tbp]
    \centering
    \caption{Experimental results of different models.}
    \label{tbl_experiments}
    \begin{tabular}{|c|c|c|c|c|c|}
    \hline
    \textbf{Model} & \textbf{F1-Score} & \textbf{mAP@.5} & \textbf{mAP@.5 : .95} & \textbf{Frame-rate (fps)} & \textbf{Inference (ms)} \\ \hline
    \textit{\yolos~(proposed)} & 0.885 & 0.918 & 0.663 & 69.4 & 14.4 \\ \hline
    \textit{Improved \ac{YOLO}v4 \cite{boyuan2020study}} & 0.80 & 0.847 & - & 36.4 & - \\ \hline
    \end{tabular}
\end{table*}

\figurename{\ref{fig_curves}} demonstrates ten curves of changes in key indicators according while training during the defined epochs. The top five curves are related to network training and the bottom five curves are related to network validation.
Here, \yolo's loss function is composed of three parts: 
\textit{$box_loss$} is a bounding box regression loss that measures how tight the predicted bounding boxes are \wrt to the ground truth object (loss criteria in use is CIoU()).
\(\textit{$obj_loss$}\) is the objectness loss that represent the confidence of object presence (Binary Cross Entropy()).
And finally, \(\textit{$cls_loss$}\) is a classification loss that measures the correctness of the classification of each predicted bounding box.
This loss is usually called Cross Entropy loss().
Since our dataset has one class only, the classification error is constantly zero.
\ac{AP} is the average over multiple \acp{IoU}, and \ac{mAP}@[.5:.95] corresponds to the average \ac{AP} for \ac{IoU} from 0.5 to 0.95 with a step size of 0.05.

In this regard, the precision-recall curve is shown in \figurename{\ref{fig_prcurve}}.
As mentioned above, the curve is obtained by plotting the model's precision and recall values as a function of the model's confidence score threshold.
The smaller the confidence threshold, the higher the recall.
On the other hand, the higher the confidence threshold, higher the precision.
Therefore, there exists a trade-off between precision and recall based on the value of the confidence threshold. 
With the precision-recall curve, we can see that \yolo~model has the high precision and recall values at the pedestrian detection task.

Table \ref{tbl_experiments} presents the accuracy of the proposed pedestrian detection system.
Accordingly, the system could achieve $0.935$ in precision and $0.84$ in the recall.
Also, the mean average precision at $50$\% \ac{IoU} (cut-off value of 5) was around $91.8$\%, which indicates something.

Moreover, to assess the system's performance, we noticed that the detection speed of the \yolos~architecture is $14.4ms$, which indicates a real-time performance.
It can also work properly for pedestrian detection when the frame rate is $~69$ fps. The rate of frame per second can increases to $277$ fps at the test when a batch size of $8$ is considered.

\subsection{Discussion}
\label{subsec_discussion}

This paper uses \yolos~as the pedestrian detector, which can meet the requirements of real-time detection.
It can be seen in Table \ref{tbl_experiments} that, compared with the other pedestrian detectors, the fine-tuned \yolos~model has obvious advantages in terms of accuracy (F1-score) and frame rate.
As a result of comparing, the \ac{mAP} of \yolos~model is $7$\% higher than \ac{YOLO}v4 \cite{boyuan2020study}.
Also, the frame rate of \yolos~is $69.4$ fps, which is approximately twice the value of \ac{YOLO}v4 ($36.4$ fps).

For further work, we plan to evaluate our system in an end-to-end application with real-world data and improve the object detection module using other robust approaches, such as transformers.
\section{Conclusions}
\label{sec_conclude}

Pedestrians are among the paramount objects that autonomous vehicles must detect.
Considering the current challenges in pedestrian detection and the prominent features introduced in \yolo, including a lower training time, auto-learning anchor bounding boxes, and mosaic data augmentation, this paper presented a pedestrian detection application based on \yolos.
The framework is written in the Ultralytics PyTorch framework, making it very fast to train.
In this approach, we have utilized the \caltech and implemented a conversion toolbox to prepare proper feed for the \ac{YOLO} architecture.
The mentioned toolbox converted images and annotations of the \caltech to items recognizable by \yolo.
According to the experiments, the mAP value of the proposed system is 91.8\%, while the frame-rate is 69.4 frames-per-second.

\section*{Acknowledgment}
The training and evaluation processes of the proposed work could not been done without the supports of the Amirkabir University of Technology for providing the \textit{Simorgh Supercomputer} under Contract No. ISI-DCE-DOD-Cloud-900808-1700.

\bibliographystyle{IEEEtran}
\bibliography{paper}

\begin{thebibliography}{10}
\providecommand{\url}[1]{#1}
\csname url@samestyle\endcsname
\providecommand{\newblock}{\relax}
\providecommand{\bibinfo}[2]{#2}
\providecommand{\BIBentrySTDinterwordspacing}{\spaceskip=0pt\relax}
\providecommand{\BIBentryALTinterwordstretchfactor}{4}
\providecommand{\BIBentryALTinterwordspacing}{\spaceskip=\fontdimen2\font plus
\BIBentryALTinterwordstretchfactor\fontdimen3\font minus
  \fontdimen4\font\relax}
\providecommand{\BIBforeignlanguage}[2]{{%
\expandafter\ifx\csname l@#1\endcsname\relax
\typeout{** WARNING: IEEEtran.bst: No hyphenation pattern has been}%
\typeout{** loaded for the language `#1'. Using the pattern for}%
\typeout{** the default language instead.}%
\else
\language=\csname l@#1\endcsname
\fi
#2}}
\providecommand{\BIBdecl}{\relax}
\BIBdecl

\bibitem{adnan2018trust}
N.~Adnan, S.~M. Nordin, M.~A. bin Bahruddin, and M.~Ali, ``How trust can drive
  forward the user acceptance to the technology? in-vehicle technology for
  autonomous vehicle,'' \emph{Transportation research part A: policy and
  practice}, vol. 118, pp. 819--836, 2018.

\bibitem{zhang2017towards}
S.~Zhang, R.~Benenson, M.~Omran, J.~Hosang, and B.~Schiele, ``Towards reaching
  human performance in pedestrian detection,'' \emph{IEEE transactions on
  pattern analysis and machine intelligence}, vol.~40, no.~4, pp. 973--986,
  2017.

\bibitem{hnewa2020object}
M.~Hnewa and H.~Radha, ``Object detection under rainy conditions for autonomous
  vehicles: A review of state-of-the-art and emerging techniques,'' \emph{IEEE
  Signal Processing Magazine}, vol.~38, no.~1, pp. 53--67, 2020.

\bibitem{dalal2005histograms}
N.~Dalal and B.~Triggs, ``Histograms of oriented gradients for human
  detection,'' in \emph{2005 IEEE computer society conference on computer
  vision and pattern recognition (CVPR'05)}, vol.~1.\hskip 1em plus 0.5em minus
  0.4em\relax Ieee, 2005, pp. 886--893.

\bibitem{lowe2004distinctive}
D.~G. Lowe, ``Distinctive image features from scale-invariant keypoints,''
  \emph{International journal of computer vision}, vol.~60, no.~2, pp. 91--110,
  2004.

\bibitem{bay2006surf}
H.~Bay, T.~Tuytelaars, and L.~V. Gool, ``Surf: Speeded up robust features,'' in
  \emph{European conference on computer vision}.\hskip 1em plus 0.5em minus
  0.4em\relax Springer, 2006, pp. 404--417.

\bibitem{viola2001rapid}
P.~Viola and M.~Jones, ``Rapid object detection using a boosted cascade of
  simple features,'' in \emph{Proceedings of the 2001 IEEE computer society
  conference on computer vision and pattern recognition. CVPR 2001},
  vol.~1.\hskip 1em plus 0.5em minus 0.4em\relax Ieee, 2001, pp. I--I.

\bibitem{da2017artificial}
I.~N. Da~Silva, D.~H. Spatti, R.~A. Flauzino, L.~H.~B. Liboni, and S.~F. dos
  Reis~Alves, ``Artificial neural networks,'' \emph{Cham: Springer
  International Publishing}, vol.~39, 2017.

\bibitem{gu2018recent}
J.~Gu, Z.~Wang, J.~Kuen, L.~Ma, A.~Shahroudy, B.~Shuai, T.~Liu, X.~Wang,
  G.~Wang, J.~Cai \emph{et~al.}, ``Recent advances in convolutional neural
  networks,'' \emph{Pattern recognition}, vol.~77, pp. 354--377, 2018.

\bibitem{jiao2019survey}
L.~Jiao, F.~Zhang, F.~Liu, S.~Yang, L.~Li, Z.~Feng, and R.~Qu, ``A survey of
  deep learning-based object detection,'' \emph{IEEE access}, vol.~7, pp.
  128\,837--128\,868, 2019.

\bibitem{redmon2016you}
J.~Redmon, S.~Divvala, R.~Girshick, and A.~Farhadi, ``You only look once:
  Unified, real-time object detection,'' in \emph{Proceedings of the IEEE
  conference on computer vision and pattern recognition}, 2016, pp. 779--788.

\bibitem{ren2015faster}
S.~Ren, K.~He, R.~Girshick, and J.~Sun, ``Faster r-cnn: Towards real-time
  object detection with region proposal networks,'' \emph{Advances in neural
  information processing systems}, vol.~28, 2015.

\bibitem{redmon2017yolo9000}
J.~Redmon and A.~Farhadi, ``Yolo9000: better, faster, stronger,'' in
  \emph{Proceedings of the IEEE conference on computer vision and pattern
  recognition}, 2017, pp. 7263--7271.

\bibitem{redmon2018yolov3}
------, ``Yolov3: An incremental improvement,'' \emph{arXiv preprint
  arXiv:1804.02767}, 2018.

\bibitem{bochkovskiy2020yolov4}
A.~Bochkovskiy, C.-Y. Wang, and H.-Y.~M. Liao, ``Yolov4: Optimal speed and
  accuracy of object detection,'' \emph{arXiv preprint arXiv:2004.10934}, 2020.

\bibitem{lin2014microsoft}
T.-Y. Lin, M.~Maire, S.~Belongie, J.~Hays, P.~Perona, D.~Ramanan,
  P.~Doll{\'a}r, and C.~L. Zitnick, ``Microsoft coco: Common objects in
  context,'' in \emph{European conference on computer vision}.\hskip 1em plus
  0.5em minus 0.4em\relax Springer, 2014, pp. 740--755.

\bibitem{jocher2021ultralytics}
G.~Jocher, A.~Stoken, J.~Borovec, A.~Chaurasia, L.~Changyu, A.~Laughing,
  A.~Hogan, J.~Hajek, L.~Diaconu, Y.~Marc \emph{et~al.}, ``ultralytics/yolov5:
  v5. 0-yolov5-p6 1280 models aws supervise. ly and youtube integrations,''
  \emph{Zenodo}, vol.~11, 2021.

\bibitem{boyuan2020study}
W.~Boyuan and W.~Muqing, ``Study on pedestrian detection based on an improved
  yolov4 algorithm,'' in \emph{2020 IEEE 6th International Conference on
  Computer and Communications (ICCC)}.\hskip 1em plus 0.5em minus 0.4em\relax
  IEEE, 2020, pp. 1198--1202.

\bibitem{tian2015deep}
Y.~Tian, P.~Luo, X.~Wang, and X.~Tang, ``Deep learning strong parts for
  pedestrian detection,'' in \emph{Proceedings of the IEEE international
  conference on computer vision}, 2015, pp. 1904--1912.

\bibitem{noh2018improving}
J.~Noh, S.~Lee, B.~Kim, and G.~Kim, ``Improving occlusion and hard negative
  handling for single-stage pedestrian detectors,'' in \emph{Proceedings of the
  IEEE conference on computer vision and pattern recognition}, 2018, pp.
  966--974.

\bibitem{zeng2014deep}
X.~Zeng, W.~Ouyang, M.~Wang, and X.~Wang, ``Deep learning of scene-specific
  classifier for pedestrian detection,'' in \emph{European Conference on
  Computer Vision}.\hskip 1em plus 0.5em minus 0.4em\relax Springer, 2014, pp.
  472--487.

\bibitem{li2019deep}
G.~Li, Y.~Yang, and X.~Qu, ``Deep learning approaches on pedestrian detection
  in hazy weather,'' \emph{IEEE Transactions on Industrial Electronics},
  vol.~67, no.~10, pp. 8889--8899, 2019.

\bibitem{liu2019high}
W.~Liu, S.~Liao, W.~Ren, W.~Hu, and Y.~Yu, ``High-level semantic feature
  detection: A new perspective for pedestrian detection,'' in \emph{Proceedings
  of the IEEE/CVF conference on computer vision and pattern recognition}, 2019,
  pp. 5187--5196.

\bibitem{ren2017accurate}
J.~Ren, X.~Chen, J.~Liu, W.~Sun, J.~Pang, Q.~Yan, Y.-W. Tai, and L.~Xu,
  ``Accurate single stage detector using recurrent rolling convolution,'' in
  \emph{Proceedings of the IEEE conference on computer vision and pattern
  recognition}, 2017, pp. 5420--5428.

\bibitem{zhang2018occluded}
S.~Zhang, J.~Yang, and B.~Schiele, ``Occluded pedestrian detection through
  guided attention in cnns,'' in \emph{Proceedings of the IEEE conference on
  Computer Vision and Pattern Recognition}, 2018, pp. 6995--7003.

\bibitem{wang2015boosting}
L.~Wang and B.~Zhang, ``Boosting-like deep learning for pedestrian detection,''
  \emph{arXiv preprint arXiv:1505.06800}, 2015.

\bibitem{tumas2020pedestrian}
P.~Tumas, A.~Nowosielski, and A.~Serackis, ``Pedestrian detection in severe
  weather conditions,'' \emph{IEEE Access}, vol.~8, pp. 62\,775--62\,784, 2020.

\bibitem{wang2020cspnet}
C.-Y. Wang, H.-Y.~M. Liao, Y.-H. Wu, P.-Y. Chen, J.-W. Hsieh, and I.-H. Yeh,
  ``Cspnet: A new backbone that can enhance learning capability of cnn,'' in
  \emph{Proceedings of the IEEE/CVF conference on computer vision and pattern
  recognition workshops}, 2020, pp. 390--391.

\bibitem{he2015spatial}
K.~He, X.~Zhang, S.~Ren, and J.~Sun, ``Spatial pyramid pooling in deep
  convolutional networks for visual recognition,'' \emph{IEEE transactions on
  pattern analysis and machine intelligence}, vol.~37, no.~9, pp. 1904--1916,
  2015.

\bibitem{liu2018path}
S.~Liu, L.~Qi, H.~Qin, J.~Shi, and J.~Jia, ``Path aggregation network for
  instance segmentation,'' in \emph{Proceedings of the IEEE conference on
  computer vision and pattern recognition}, 2018, pp. 8759--8768.

\bibitem{dollar2011pedestrian}
P.~Dollar, C.~Wojek, B.~Schiele, and P.~Perona, ``Pedestrian detection: An
  evaluation of the state of the art,'' \emph{IEEE transactions on pattern
  analysis and machine intelligence}, vol.~34, no.~4, pp. 743--761, 2011.

\end{thebibliography}

\end{document}